# Large-scale pretraining on pathological images for fine-tuning of small pathological benchmarks

Masakata Kawai [1][0000-0003-1106-239X] Noriaki Ota [2] Shinsuke Yamaoka [2]

[1] University of Yamanashi, Department of Pathology, Japan
mkawai@yamanashi.as.jp
[2] Systems Research & Development Center, Technology Bureau, NS Solutions Corp., Japan
{ota.noriaki.4qp, yamaoka.shinsuke.5ke}@jp.nssol.nipponsteel.com

**Abstract.** Pretraining a deep learning model on large image datasets is a standard step before fine-tuning the model on small targeted datasets. The large dataset is usually general images (e.g. imagenet2012) while the small dataset can be specialized datasets that have different distributions from the large dataset. However, this "large-to-small" strategy is not well-validated when the large dataset is specialized and has a similar distribution to small datasets. We newly compiled three hematoxylin and eosin-stained image datasets, one large (**PTCGA200**) and two magnification-adjusted small datasets (**PCam200** and **segPANDA200**). Major deep learning models were trained with supervised and self-supervised learning methods and fine-tuned on the small datasets for tumor classification and tissue segmentation benchmarks. ResNet50 pretrained with MoCov2, SimCLR, and BYOL on **PTCGA200** was better than imagenet2012 pretraining when fine-tuned on **PTCGA200** (accuracy of 83.94%, 86.41%, 84.91%, and 82.72%, respectively). ResNet50 pretrained on PTCGA200 with MoCov2 exceeded the CO-COtrain2017-pretrained baseline and was the best in ResNet50 for the tissue segmentation benchmark (mIoU of 63.53% and 63.22%). We found re-training imagenet-pretrained models (ResNet50, BiT-M-R50x1, and ViT-S/16) on **PTCGA200** improved downstream benchmarks.

**Keywords:** Large-scale training, Pathological benchmark, Self-supervised learning, Vision transformer.

## 1 Introduction

Large-scale pretraining benefits large neural nets for fine-tuning downstream tasks. One success story is imagenet (including imagenet2012 and imagenet-21k, or in21k) [12] for supervised and self-supervised learning of deep neural models in computer vision [1, 5, 8, 9, 18, 19, 21, 26, 34, 35, 37, 38]. Pretraining on large datasets before fine-tuning on small datasets, we call this "large-to-small" strategy, is a common strategy in deep learning [2, 14, 16, 24, 33]. However, the effect of pretraining on large specialized image datasets with different distributions from general images is not well systematically compared due to the lack of scale adjusted large and small datasets. Here, we conducted a large-scale supervised and self-supervised pretraining of deep learning models on pathological hematoxylin and eosin (H&E) stained images. After pretraining, the models were fine-tuned on small magnification-adjusted pathological datasets, benchmarking the downstream classification and semantic segmentation performances (Fig. 1).

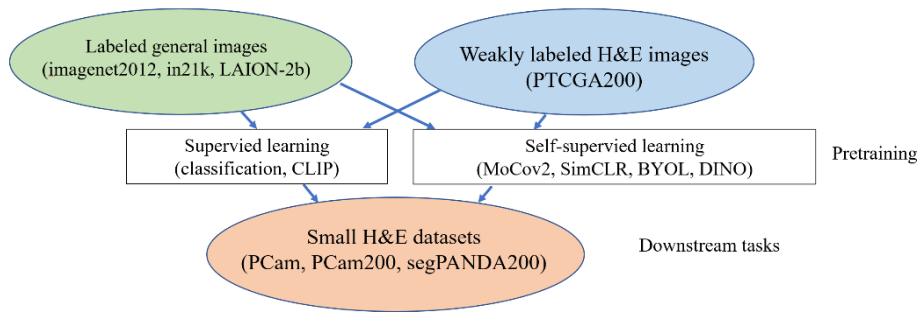

**Fig. 1.** Upstream large-scale supervised and self-supervised pretraining on general image datasets and a a H&E image dataset (**PTCGA200**). The model performance of classification and tissue segmentation was measured on the downstream small H&E image datasets.

Our findings and contributions are as follows:

- We compiled three public pathological H&E image datasets, one large (**PTCGA200**) and two small datasets (**PCam200** and **segPANDA200**), suitable for deep learning pretraining and benchmarking.
- Self-supervised learning frameworks benefit from **PTCGA200** pretraining and often exceed imagenet2012 pretraining, but not outperform in21k pretraining.
- Re-training on **PTCGA200** of imagenet2012/in21k-pretrained models often boosts the downstream task performances.

## 2    Related work

Several studies [1, 26, 33] showed large models pretrained on large image datasets are better in the downstream performance than small models and small datasets pairs. However, specialized image domains including H&E images in digital pathology are distinct from general images e.g. cats and dogs. Particularly, contrastive self-supervised learning [2, 6, 8, 9, 10, 18, 21] leveraging content-preserving augmentations needs care for these image domain differences. The success of transfer learning from general images to hematoxylin and eosin (H&E) images [10, 24], although it is a common approach, is not obvious [28]. As such, we were motivated to fill the gap of the "large-to-small strategy" in digital pathology, a large H&E dataset to small H&E datasets. We compiled a large patch-based H&E image dataset ourselves as there are no public large H&E datasets comparable with imagenet2012. Instead, several studies [2, 7, 22, 31] used a large repository of whole slide H&E images (WSIs) from the Cancer Genome Atlas (TCGA) project. However, WSIs are gigapixel images including many uninformative white backgrounds. A patch-based dataset is handy, compact, and memory efficient in the deep learning pipeline and improves reproducibility. For the above reasons, we newly introduced a large patch-based dataset cropped from TCGA WSIs named **PTCGA200** that is comparable with imagenet2012. Model performance is often measured by downstream task performance using learning visual task adaptation benchmark (VTAB) [43], a suit of 19 distinct small datasets to benchmark transfer learning capability [1, 26, 33]. VTAB includes a pathological dataset, Patch Camelyon (denotes **PCam**) [15, 40], as one of the specialized datasets. However, the 96px images at 0.24 microns per pixel (MPP) in **PCam** are small for the recent computational environment and hard to adjust to the magnification of **PTCGA200**. To benchmark classification and segmentation performances, two small magnification-adjusted patch-based datasets (**PCam200** and **segPANDA200**) were newly introduced. Newly compiled datasets as well as the most of the experimental codes will be public[1].

## 3    Method

We benchmarked supervised and self-supervised pretraining on a large pathological image dataset and validated the fine-tuning performance on downstream pathological tasks. We randomly cropped 500 patches in the 200μm scale per slide from the tissue region of around 10220 diagnostic slides from TCGA and resized them into $512{\times}512$px using bicubic interpolation. Training, validation, and test sets were split so that patches from the same slide be in the same set (slide-level split). The RGB means and standard deviations of the whole **PTCGA200** are (0.7184, 0.5076, 0.6476) and (0.0380, 0.0527, 0.0352), respectively. **PCam200** was made in the same manner from Camelyon2016 challenge dataset [15]. **SegPANDA200** was made in the same manner from PANDA challenge dataset [3], but the patch size was 1024px at the same MPP as **PTCGA200**. Training, validation, and test sets were split so that the proportion of ISUP grade is balanced in each set. The supervised pretraining task on **PTCGA200** is a 20-class organ classification, while the downstream tasks are binary tumor or normal classification (**PCam200** and **PCam**), and 6-class semantic segmentation of prostate biopsy (**segPANDA200**). They are in the same H&E image domain but, by definition, are distinct and almost independent tasks of each other. The datasets used in this study are tabulated in Table 1. The details of dataset compilation are found in Appendix.

**Table 1.** Properties of the datasets used in this study. MPP: microns per pixel.

| Dataset | Clas-ses | Training | Valida-tion | Test | Image size | MPP |
| --- | --- | --- | --- | --- | --- | --- |

---

| | | | | | | |
|---|---|---|---|---|---|---|
| imagenet2012 | 1000 | 1281167 | 50000 | 100000 | Variable | - |
| in21k | 21841 | Total 14197122 | | | Variable | - |
| LAION-2b | - | Total 2b | | | Variable | - |
| PTCGA200 | 20 | 4945500 | 107500 | 57000 | 512 | 0.39 |
| PCam200 | 2 | 28539 | 10490 | 17674 | 512 | 0.39 |
| PCam | 2 | 262144 | 32768 | 32768 | 96 | 0.97 |
| segPANDA200 | 6 | 70878 | 15042 | 15040 | 1024 | 0.39 |

The image augmentation suits described in Appendix were applied to absorb the color differences in different datasets. The experiments were conducted almost only once except for a minor hyperparameter search on validation sets (up to 5 trials). The experiments were conducted with multiple NVIDIA V100 GPUs except for ViT-S/16 supervised scratch training with two A6000 NVIDIA GPUs. PyTorch and torchvision versions were 1.7.1 and 0.8.2, respectively.

**Fine-tuning details.** We replaced the classification head with a D × N fully connected layer in **PTCGA200**, **PCam200**, and **PCam**, where D is the dimension of the encoder output and N is the number of classes. We reset the running means and variances in the batch normalization layers before fine-tuning. We adopted the same hyperparameters in each task, prioritizing model varieties over better performance by hyperparameter search. The optimizer was SGD with momentum=0.9 and Nesterov=False without weight decay. The batch size was 512. The learning rate was 0.05 and decayed according to the cosine annealing schedule. The images were normalized using the same means and standard deviations as pretraining. The RGB means and standard deviations of (0.5, 0.5, 0.5) and (0.5, 0.5, 0.5) respectively were used when fine-tuning from randomly initialized weights. Some models were fine-tuned on larger input images (384px) than pretraining as recommended in [1, 26]. The positional embedding of ViT was resized to the targeted size by bicubic interpolation. Linear classification protocols [8, 18, 21, 43] are often used to evaluate the backbone model. Nevertheless, we fine-tuned the whole model to simulate real applications requiring better overall performance.

## 4 Experiment

### 4.1 PTCGA200 scratch pretraining

We trained ResNet18/50 [19], InceptionV3 [37], EfficientNet-b3 [38], ViT-S/16 [1], and ViT-B/32 [1] as the deep learning models for scratch pretraining. We excluded VGG-16 [35] although they are popular in digital pathology applications. They lack batch normalization layers [23] and have nearly five times the number of parameters as compared to ResNet50 (130M versus 22M, respectively), so the direct comparison seemed inappropriate. We included vision transformers (ViTs) [1] as recent studies report their utility in digital pathology [7, 12, 27]. All the models were trained on **PTCGA200** for the 20-class classification task minimizing cross-entropy loss. The weights were randomly initialized. The images were normalized using **PTCGA200** means and standard deviations. The optimizer was AdamW [30] with beta1=0.9 and beta2=0.999. The batch size was 4k. The learning rates were increased in the warmup epochs [17] and decayed with the cosine annealing schedule in the remaining epochs. The base learning rate was increased during warmup epochs until the peak learning rate was equal to the base learning rate multiplied by (batch size)/256. We report top-1 accuracies in % in Table 2.

**Table 2.** Top-1 accuracies (in %) supervised scratch training on **PTCGA200**.

| Model | Test accuracy | Best validation accuracy | Epochs | Image size | Weight decay | Base LR | Warmup epochs |
|---|---|---|---|---|---|---|---|
| ResNet18 | 84.01 | 83.94 | 60 | 224 | 5e-5 | 0.001 | 10 |
| ResNet50 | 85.30 | 85.32 | 60 | 224 | 5e-5 | 5e-4 | 10 |
| InceptionV3 | 85.98 | 85.66 | 60 | 299 | 5e-5 | 1e-4 | 10 |
| EfficientNet-b3 | **88.35** | 88.11 | 60 | 300 | 5e-5 | 1e-4 | 10 |
| ViT-S/16 | 84.44 | 86.63 | 80 | 224 | 0.03 | 1e-4 | 15 |
| ViT-B/32 | 84.10 | 84.83 | 100 | 224 | 0.03 | 1e-4 | 20 |

All the models obtained over 84.01% accuracy. EfficientNet-b3 obtained the best accuracy of 88.35%, which was the best among all the models including the following fine-tuning part.

**Self-supervised scratch pretraining.** We included self-supervised learning as it recently attracts attention [7, 12, 39] in digital pathology where detailed patch-level labels are hard to obtain. We chose MoCov2 [21], SimCLR [8], BYOL [18], and DINO [5] as the self-supervised methods. They seek invariant representation across two differently augmented views encoded by a pair of slightly different encoders [4]. The backbone models were Res-Net50 in SimCLR, MoCov2, and BYOL, and ViT-S/16 in DINO. We trained Mo-Cov2, SimCLR, BYOL, and DINO on PTCGA200 training set from scratch for 40 epochs. We modified SimCLRv1 [8] to have 3-layer MLP and made the first layer of the MLP the encoder output as proposed in SimCLRv2 [9]. We implemented BYOL in PyTorch following the original implementation in JAX. We froze patch embedding of ViT in DINO as in [10]. The images were normalized using PTCGA200 means and standard deviations. The hyperparameters were set default in the original repositories.

### 4.2 PTCGA200 fine-tuning of pretrained models

We fine-tuned imagenet2012-, in21k-, and LAION-2b-pretrained models downloaded from PyTorch Hub, PyTorch Image Models [41], or official repositories (re-training) as well as **PTCGA200**-pretrained models on **PTCGA200** for 30k iterations. Supervised re-training was inspired by the success of self-supervised re-training in REMEDIS [2]. The downloaded ResNet50-McCov2 were pretrained on imagenet2012 for 300epochs. We included Big Transfer (BiT) [26] models pretrained on imagenet2012 and/or in21k (BiT-S and BiT-M respectively in the original paper) as the state-of-the-art models for transfer learning. BiT-R50x1 has a ResNetv2 [20] architecture characterized by group normalization [42] and weight standardization [32] instead of batch normalization. We report top-1 accuracies in % in Tables 3 and 4.

**Table 3.** PTCGA200 top-1 accuracies (in %) fine-tuned from supervised pretraining on general images.

| Model | Test accuracy | Image size | Pretraining dataset |
|---|---|---|---|
| ResNet18 | 79.27 / 81.79 | 224 / 384 | imagenet2012 |
| ResNet50 | 81.46 / 82.72 | 224 / 384 | imagenet2012 |
| InceptionV3 | 83.05 | 384 | imagenet2012 |
| EfficientNet-b3 | 83.39 | 384 | imagenet2012 |
| ViT-S/16 | 85.13 / **87.21** | 224 / 384 | in21k |
| ViT-B/32 | 82.36 / 85.87 | 224 / 384 | in21k |
| ViT-B/32-CLIP | 66.06 / 75.13 | 224 / 384 | LAION-2b |
| BiT-R50x1 | 85.64 | 384 | in21k |

**Table 4.** PTCGA200 top-1 accuracies (in %) fine-tuned from self-supervised pretraining

| Model | Test accuracy | Image size | Pretraining dataset |
|---|---|---|---|
| ResNet50-McCov2 | 81.06 / 83.09 | 224 / 384 | imagenet2012 |
| ResNet50-McCov2 | 82.83 / 83.94 | 224 / 384 | PTCGA200 |
| ResNet50-SimCLR | 86.31 / **86.41** | 224 / 384 | PTCGA200 |
| ResNet50-BYOL | 83.73 / 84.91 | 224 / 384 | PTCGA200 |
| ViT-S/16-DINO | 69.09 / 72.97 | 224 / 384 | imagenet2012 |
| ViT-S/16-DINO | 83.60 / 85.05 | 224 / 384 | PTCGA200 |

ViT-S/16 pretrained on in21k obtained the best performance of 87.21% accuracy in fine-tuned models. **PTCGA200**-pretrained ResNet50 with SimCLR fine-tuned on 384px images was the best in ResNet50 including scratch training. Fine-tuned ViT-S/16, ViT-B/32, and BiT-R50x1 on 384px images as well as ResNet50 pretrained on PTCGA200 with SimCLR exceeded the ResNet50 scratch training baseline of 85.30%. Fine-tuning on 384px images improved the performance invariably. **PTCGA200**-pretrained self-supervised models (MoCov2 and DINO) exceeded the corresponding imagenet2012-pretrained models.

### 4.3 PCam200 fine-tuning

We fined-tuned models used in PTCGA200 on PCam200 for 1k iterations. We report accuracies in % in Tables 5 and 6.

**Table 5.** **PCam200** accuracies (in %) fine-tuned from supervised pretraining on **PTCGA200** and general images. Pretraining dataset **None** indicates random initialization of weights. The right side of the arrow indicates the re-training dataset.

| Model | Test accuracy | Image size | Pretraining dataset |
|---|---|---|---|

| ResNet18 | 92.72 / 82.14 | 384 | imagenet2012 / PTCGA200 |
| ResNet50 | 79.10 | 224 | None |
| ResNet50 | 91.97 / 92.56 | 224 / 384 | imagenet2012 |
| ResNet50 | 91.89 / 92.48 | 224 / 384 | PTCGA200 |
| ResNet50 | **93.24** | 384 | imagenet2012→PTCGA200 |
| InceptionV3 | 92.57 / 91.80 | 384 | imagenet2012 / PTCGA200 |
| EfficientNet-b3 | 90.51 / 89.69 | 384 | imagenet2012 / PTCGA200 |
| ViT-S/16 | 92.91 / 88.80 | 384 | in21k / PTCGA200 |
| ViT-S/16 | 93.22 | 384 | in21k→PTCGA200 |
| ViT-B/32 | 91.65 / 89.62 | 384 | in21k / PTCGA200 |
| ViT-B/32-CLIP | 76.75 | 384 | LAION-2b |
| BiT-R50x1 | 93.24 / **93.67** | 384 | in21k→imagenet2012/in21k |
| BiT-R50x1 | 93.49 | 384 | in21k→PTCGA200 |

**Table 6.** PCam200 accuracies (in %) fine-tuned from self-supervised pretraining on **PTCGA200** and general images.

| Model | Test accuracy | Image size | Pretraining dataset |
|---|---|---|---|
| ResNet50-MoCov2 | 91.70 / 91.94 | 384 | imagenet2012 / PTCGA200 |
| ResNet50-SimCLR | 92.07 | 384 | PTCGA200 |
| ResNet50-BYOL | **92.15** | 384 | PTCGA200 |
| ViT-S/16-DINO | 76.16 / 91.79 | 384 | imagenet2012 / PTCGA200 |

BiT-R50x1 pretrained on in21k obtained the best accuracy of 93.67%. Re-training of imaget2012/in21k-pretrained models on **PTCGA200** improved the accuracies by 0.68% (ResNet50) and 0.25% (BiT-R50x1). Re-training on **PTCGA200** from imagenet2012-pretraining was the best in ResNet50 with an accuracy of 93.24%. Imagenet2012/in21k-pretrained models exceeded **PTCGA200**-pretrained supervised models, but competing in ResNet50 (92.56% and 92.48%), InceptionV3 (92.57% and 91.80%) and EfficientNet-b3 (90.51% and 89.69%). Fine-tuning on 384px images improved the performance. **PTCGA200**-pretrained self-supervised models (MoCov2 and DINO) exceeded the corresponding imagenet2012-pretrained models (91.94% and 91.70%, and 91.79% and 76.16%, respectively).

### 4.4 PCam fine-tuning

We fined-tuned models used in **PTCGA200** on **PCam** [15, 40] for 1k iterations. We report accuracies in % in Tables 7 and 8. Reference performances were cited from the published papers [33, 43].

**Table 7.** PCam accuracies (in %) fine-tuned from supervised pretraining on **PTCGA200** and general images. Pretraining dataset **None** indicates random initialization of weights. The right side of the arrow indicates the re-training dataset.

| Model | Test accuracy | Image size | Pretraining dataset |
|---|---|---|---|
| ResNet18 | 87.55 | 384 | imagenet2012 |
| ResNet18 | 81.57 / 80.22 | 224 / 384 | PTCGA200 |

| ResNet50 | 79.10 / 79.03 | 224 / 384 | None |
|---|---|---|---|
| ResNet50 | 89.16 / 89.05 | 224 / 384 | imagenet2012 |
| ResNet50 | 79.13 / 76.03 | 224 / 384 | PTCGA200 |
| ResNet50 | 87.44 / 87.05 | 224 / 384 | imagenet2012→PTCGA200 |
| ResNet50 | 87.3 [42] | 224 | imagenet2012 |
| ResNet50 | **91.2** [42] | 224 | None |
| InceptionV3 | 87.93 | 384 | imagenet2012 |
| InceptionV3 | 86.49 | 384 | PTCGA200 |
| EfficientNet-b3 | 88.23 | 384 | imagenet2012 |
| EfficientNet-b3 | 82.59 | 384 | PTCGA200 |
| ViT-S/16 | 90.16 / **90.51** | 224 / 384 | in21k |
| ViT-S/16 | 84.35 / 85.29 | 224 / 384 | PTCGA200 |
| ViT-S/16 | 88.08 / 89.87 | 224 / 384 | in21k→PTCGA200 |
| ViT-B/32 | 89.24 | 384 | in21k |
| ViT-B/32 | 86.05 / 85.89 | 224 / 384 | PTCGA200 |
| ViT-B/32-CLIP | 78.60 | 384 | LAION-2b |
| ViT-B/32-CLIP | 82.6 [32] | 224 | WebImageText [32] |
| BiT-R50x1 | 88.43 / 88.25 | 224 / 384 | in21k |
| BiT-R50x1 | 86.33 | 384 | in21k→PTCGA200 |
| BiT-R50x1 | 87.67 / 87.38 | 224 / 384 | in21k |

**Table 8. PCam** accuracies (in %) fine-tuned from self-supervised pretraining on **PTCGA200** and general images.

| Model | Test accuracy | Image size | Pretraining dataset |
|---|---|---|---|
| ResNet50-MoCov2 | 87.94 / 88.97 | 224 / 384 | imagenet2012 |
| ResNet50-MoCov2 | 89.23 / 89.38 | 224 / 384 | PTCGA200 |
| ResNet50-SimCLR | 85.32 / 80.90 | 224 / 384 | PTCGA200 |
| ResNet50-BYOL | **89.90** / 88.95 | 224 / 384 | PTCGA200 |
| ViT-S/16-DINO | 77.42 | 384 | imagenet2012 |
| ViT-S/16-DINO | 89.44 / 88.95 | 224 / 384 | PTCGA200 |

ViT-S/16 pretrained on in21k obtained the best accuracy of 90.51%. Retraining of imagnet2012 or in21k pretrained models on **PTCGA200** degraded the accuracies by 0.64 to 2.08%. Fine-tuning on 384px images degraded the performance except for ViT-S/16 pretrained on in21k. **PTCGA200**-pretrained self-supervised ResNet50 (MoCov2 and BYOL) fine-tuned on 224px images exceeded the corresponding imagenet2012-pretrained supervised model accuracy of 89.16%. **PCam** results were similar to PCam200 results although the dataset compilation processed were totally different. Notably, **PCam** fine-tuning on 384px images (at MPP of 0.24) often degrade accuracies imagenet2012/in21k-pretrained models as well as **PTCGA200**-pretrained models. This may be because MPP of 0.24 or 40x equivalent magnification is inappropriate for the task.

#### 4.5 SegPANDA200 fine-tuning

We fined-tuned models used in **PTCGA200** fine-tuning as well as COCOtrain2017-pretrained models on **segPANDA200** for 1k iterations. In **segPANDA200** a fully convolutional network (FCN) [29] head was attached to the feature maps or the ViT maps defined as the following. ViT is a convolution-free network and has no feature maps, so we retiled the output sequence excluding **CLS** token as the original patch positions. This is the reverse operation of flattening before patch embedding. We call this feature map like intermediate output the ViT map. We used the last layer (12th layer of ViT-S and ViT-B) output and zero-initialized the positional embedding. We examined which layer output to use and the effect of zero-initializing positional encoding (not shown) and found the last layer output is good when the positional embedding is zero-initialized. We report mIoU (mean intersection over union) in % in Tables 9 and 10.

**Table 9.** SegPANDA200 mIoU (in %) fine-tuned from supervised pretraining on **PTCGA200** and general images. Pretraining dataset **None** indicates random initialization of weights. The right side of the arrow indicates the re-training dataset. Feature/ViT map size is intermediated output size in pixels on the validation/test 1024px image before FCN or DeepLabv3 heads [6].

| Model | Test mIoU | Pretraining dataset | Feature/ViT map size |
|---|---|---|---|
| ResNet18 | 57.76 / 48.71 | imagenet2012 / PTCGA200 | 32 |
| ResNet50 | 43.27 | None | 32 |
| ResNet50 | 58.76 / 49.79 | imagenet2012 / PTCGA200 | 32 |
| ResNet50 | 59.93 | imagenet2012→PTCGA200 | 32 |
| InceptionV3 | 58.70 / 50.12 | imagenet2012 / PTCGA200 | 30 |
| EfficientNet-b3 | 61.31 / 51.42 | imagenet2012 / PTCGA200 | 32 |
| ViT-S/16 | 61.57 / 57.70 | in21k / PTCGA200 | 64 |
| ViT-S/16 | 61.23 | in21k→PTCGA200 | 64 |
| ViT-B/32 | 59.21 / 58.04 | in21k / PTCGA200 | 32 |
| ViT-B/32-CLIP | 40.08 | LAION-2b | 32 |
| BiT-R50x1 | 67.16 / 67.07 | in21k→imagenet2012/in21k | 32 |
| BiT-R50x1 | **67.60** | in21k→PTCGA200 | 32 |
| ResNet50 | 63.22 | COCOtrain2017 (PASCAL VOC) | 32 |
| ResNet50-DeepLabv3 | **69.61** | COCOtrain2017 (PASCAL VOC) | 32 |

**Table 10.** SegPANDA200 mIoU (in %) fine-tuned from self-supervised pretraining on **PTCGA200** and general images. Feature/ViT map size is intermediated output size in pixels on the validation/test 1024px image before FCN heads.

| Model | Test mIoU | Pretraining dataset | Feature/ViT map size |
|---|---|---|---|
| ResNet50-MoCov2 | 62.76 / **63.53** | imagenet2012 / PTCGA200 | 32 |
| ResNet50-SimCLR | 53.50 | PTCGA200 | 32 |

| ResNet50-BYOL | 56.01 | PTCGA200 | 32 |
| ViT-S/16-DINO | 36.21 / 59.26 | imagenet2012 / PTCGA200 | 64 |

In21k-pretrained BiT-R50x1 re-trained on **PTCGA200** obtained the best mIoU of 67.60%. Re-training of imaget2012 pretrained models on **PTCGA200** improved the mIoUs by 1.17% (ResNet50) and 0.44% (BiT-R50x1). But it degraded the mIoU by 0.34% in ViT-S/16. **PTCGA200**-pretrained self-supervised ResNet50 with MoCov2 obtained the best mIoU of 63.53% in ResNet50 and exceeded the imagenet2012-pretrained mIoU of 58.76% and COCOtrain2017-pretrained mIoU of 63.22%.

## 5  Conclusion

We introduced 3 novel datasets for H&E image pretraining and benchmarking. We showed large-scale training on **PTCGA200** benefits not only self-supervised models trained from scratch but also imagenet2012/in21k-pretrained models by re-training. Magnification-adjusted datasets (**PCam200** and **segPANDA**) strengthened benchmark reliability for classification and segmentation.

**Acknowledgment.** The results shown here are in part based upon data generated by the TCGA Research Network: https://www.cancer.gov/tcga. We follow the original licenses to share the compiled datasets. We share **PTCGA200** by acknowledging NIH Genomic Data Sharing (GDS) Policy. We share **PCam200** dataset under CC0 license. We share **segPANDA200** dataset under CC BY-SA-NC 4.0 license. We appreciate to L.M.Khang for reviewing the manuscript and giving precious comments.

**Funding.** JPNP20006, commissioned by the New Energy and Industrial Technology Development Organization (NEDO).
.

# 6    Appendix

## 6.1    Compiling pathological datasets

**PTCGA200** We gathered around 10000 H&E diagnostic slides from The Cancer Genome Atlas (TCGA). We randomly cropped 500 square patches in the 200μm scale per slide from the tissue region and resized them into 512×512px using bicubic interpolation. The magnification of the patch is 0.39μm per pixel (MPP). Total 5.11M images from 10,220 slides were randomly split into training, validation, and test sets so that patches from the same slide be in the same set (slide-level split). Each contained 4945500, 107500, and 57000 images, respectively. We call this dataset **PTCGA200** (Patch TCGA in the 200μm scale at 512px). We combined similar organs and defined a 20-class organ classification task as the supervised objective on **PTCGA200** (Fig. 3). The class distribution of the dataset is imbalanced with the ratio of most disproportionate classes reaching (lymph_node/brain≈0.015).

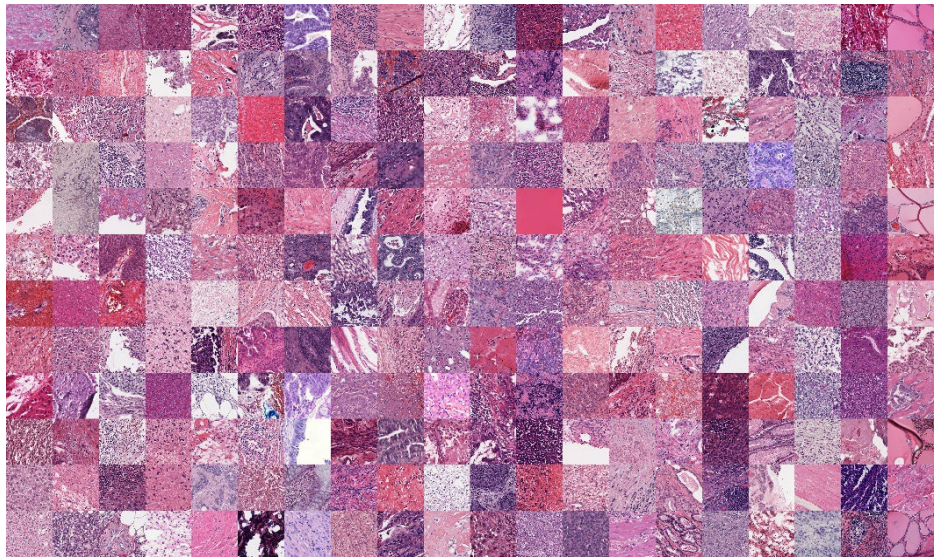

**Fig. 2.** Uncurated **PTCGA200** samples. Each column contains the patches in the same class.

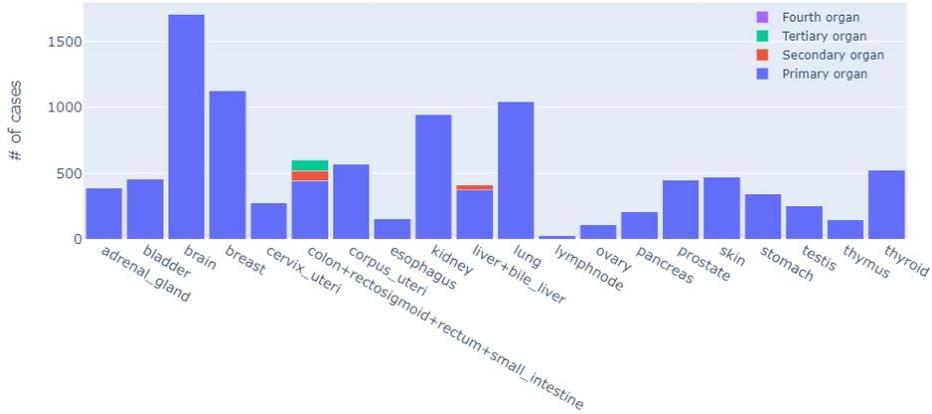

**Fig. 3.** Organ distribution of **PTCGA200**. Large and small intestine slides were united as a single intestine class.

**PCam200** We cropped square patches in the 200μm scale per slide from Camelyon2016 challenge dataset by sliding a cropping area systematically allowing some overlaps. Tumor patches were cropped from the annotated region in the tumor slides. Normal patches were cropped from the tissue region in the normal slides. Cropped patches were resized into 512×512 px using bicubic interpolation. The magnification of the patch is the same as **PTCGA200**. To match the numbers of tumor and normal patches, normal patches were randomly discarded. Total 56703 images are randomly split into training, validation, and test sets so that the patches from the original training slides go into the training or validation set and the patches from the original testing slides go into the test set. The patches from the same slide are in the same set (slide-level split). Each contained 28539, 10490, and 17674 images, respectively. We call this dataset PatchCamelyon200, or **PCam200** (Patch Camelyon2016 in the 200μm scale at 512px).

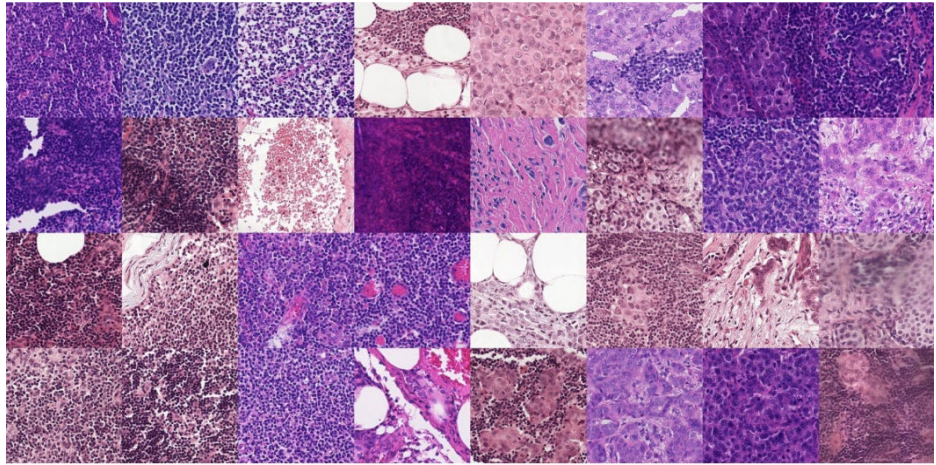

**Fig. 4.** Uncurated **PCam200** samples. The left four columns are normal patches and the right four columns are tumor patches.

**SegPANDA200** We cropped square patches in the 400µm scale per slide from PANDA challenge dataset by sliding a cropping area systematically allowing some overlaps. Slides with the data provider property of Radboud were enrolled as the segmentation masks were more detailed than slides from Karolinska. Cropped patches were resized into 1024×1024 px using bicubic interpolation. The corresponding mask images were saved as PNG images. The mask labels were the same as the original, i.e. 0: background (non tissue) or unknown, 1: stroma (connective tissue, non-epithelium tissue), 2: healthy (benign) epithelium, 3: cancerous epithelium (Gleason 3), 4: cancerous epithelium (Gleason 4), and 5: cancerous epithelium (Gleason 5). The magnification of the patch is the same as PTCGA200. Total 100960 images are randomly split into training, validation, and test sets so that the proportion of ISUP grade is balanced in each set and the patches from the same slide are in the same set (slide-level split). Each contained 70878, 15042, and 15040 images, respectively. We call this dataset **segPANDA200** (segmentation PANDA in the 200µm scale at 512px).

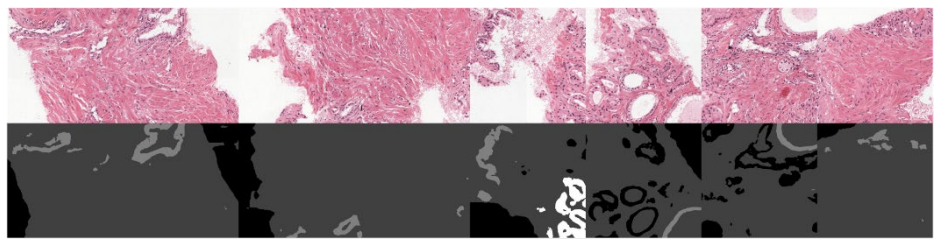

**Fig. 5.** Uncurated **segPANDA200** samples. The upper row is the original tissue images and the lower row is the corresponding segmentation masks.

### 6.2 Training miscellaneous

**Image augmentation.** The same image augmentation suit in Table 11 was applied in the training phase. Augmentation methods are from torchvision.transforms package except for GaussianBlur which is MoCov2 implementation. During validation and testing, no image augmentation was applied except for PTCGA200 and PCam200 where CenterCrop(287) was applied before RGB normalization.

**Table 11.** Augmentation suit used in training. Each method is applied sequentially from top to bottom in the list with the corresponding probabilities. RandomCrop(512) was applied in segPANDA200 instead of RandomResizedCrop. †: set 0.8 in PCam training, *: only applied in MoCoV2, SimCLR, BYOL, and DINO training.

| Augmentation methods | Probability |
|---|---|
| RandomResizedCrop(size=image_size, scale=(0.2†, 1.), ratio=(0.75, 1.3333333333333333)) | 1.0 |
| ColorJitter(brightness=0.4, contrast=0.4, saturation=0.4, hue=0.1) | 0.8 |
| RandomGrayscale* | 0.2 |
| GaussianBlur(min_sigma=0.1, max_sigma=2.0) | 0.5 |
| RandomHorizontalFlip | 0.5 |
| RandomVerticalFlip | 0.5 |

### 6.3 Magnification dependence.

Pathologists think pathological images are magnification dependent. High magnification e.g. 40x ($\approx$0.25 MPP) captures nuclear features but loses tissue architecture, while low magnification e.g. 10x ($\approx$1.0 MPP) encompasses tissue architecture but loses nuclear features. First, we changed the validation/testing crop size (Table 12). Next, we changed the magnification during training and validation/testing. Imangenet2012-preteind BiT-R50x1 was fine-tuned on **PTCGA200** using 384px images (Table 13). We report top-1 accuracies in % in Tables 12 and 13.

**Table 12.** Validation/testing crop size dependence. MPP at 384px is noted.

| Validation/testing crop size | MPP | Test accuracy |
|---|---|---|
| 512 | 0.52 | **87.94** |
| 393 | 0.40 | 87.34 |
| 287 | 0.29 | 85.64 |
| 197 | 0.20 | 81.65 |
| 98 | 0.10 | 28.34 |

**Table 13.** Training and validation/testing magnification dependence.

| Training MPP | Validation/testing MPP | Test accuracy | Accuracy change |
|---|---|---|---|
| 0.10-0.52 | 0.29 | 85.64 | - |
| 0.29 | 0.29 | 86.32 | +0.68 |
| 0.52 | 0.29 | 44.00 | -41.64 |
| 0.29 | 0.52 | 56.35 | -29.29 |

The accuracies were in parallel with the validation/testing crop sizes. This indicates the model more easily finds classification clues as it sees a larger area. The accuracy dropped dramatically when tested on MPP of 0.10. This is because MPP of 0.10 at 384px ($\approx$100x) is too small to guess the organ origin. The accuracy improved when validation/testing MPP was the same as training MPP. However, the accuracies dropped dramatically when they differ.

### 6.4 Effects of image augmentation: ablation study.

We ablated one of the augmentation methods during training to show the effect of each augmentation on the validation/testing performance. Imangenet2012-preteind BiT-R50x1 was fine-tuned on **PTCGA200**. We report top-1 accuracies in % in Table 14. The accuracy slightly improved by removing any of ColorJitter, GaussianBlur, or flipping. Given the nature of pathological images, it is counter-intuitive that removing flipping augmentation improved performance.

**Table 14.** Augmentation ablation study.

| Ablated augmentation | Test accuracy | Accuracy change |
|---|---|---|
| None (base line) | 85.64 | - |
| ColorJitter | 87.81 | +2.17 |
| GaussianBlur | 86.15 | +0.51 |
| Horizontal and vertical flip | 87.22 | +1.58 |

### 6.5 ViT map segmentation experiments.

We defined the ViT map as the retiled output sequence except **CLS** token. To see which layer's output most contributed to the segmentation performance, ViT maps of 1, 3, 5, 7, 9, 11, and 12th (default) layer output of in21k-pretrained ViT-S/16 were compared. The positional embedding was not zero-initialized. In21k-preteind ViT-S/16 was fine-tuned on **segPANDA200**. We report mIoU in % in Table 15. The mIoUs increased as the number of layers increased until the 7th layer and slightly decreased afterward.

**Table 15.** SegPANDA200 mIoU of different layer output and deviations from the default last 12th layer output.

| Output layer | Test mIoU | mIoU change |
|---|---|---|
| 1 | 44.32 | -9.15 |
| 3 | 52.90 | -0.25 |
| 5 | 53.22 | -0.25 |
| 7 | 54.15 | +0.68 |
| 9 | 53.78 | +0.31 |
| 11 | 53.70 | +0.23 |
| 12 | 53.47 | - |

Next, we compared pretrained positional embedding and zero-initialized positional embedding. We report mIoU change in % in Table 16. The mIoU jumped in in21k-pretrained ViT-S/16 from 53.47% to 61.57% and the 12th layer output result was better than the 7th layer output result of 60.19%. However, zero-initialization of the positional embedding sometimes slightly degraded the mIoUs.

**Table 16.** SegPANDA200 mIoU deviations by zero-initialization of the positional embedding.

| Model | Test mIoU change | Pretraining dataset |
|---|---|---|
| ViT-S/16 (7th layer output) | +6.04 | in21k |
| ViT-S/16 (12th layer output) | +8.10 | in21k |
| ViT-S/16 | +0.07 | PTCGA200 |
| ViT-S/16 | +1.54 | in21k→PTCGA200 |
| ViT-S/16-DINO | -1.15 | imagenet2012 |
| ViT-S/16-DINO | -0.72 | PTCGA200 |
| ViT-B/32 | -0.39 | PTCGA200 |
| ViT-B/32 | +0.10 | in21k |
| ViT-B/32-CLIP | -0.13 | LAION-2b |

### 6.6    1% PTCGA200 fine-tuning of pretrained models.

We conducted the fine-tuning experiments of 1% (slide-base split) of **PTCGA200** as in [7, 17]. We fine-tuned for 30k iterations. We report top-1 accuracies in % in Table 17.

**Table 17.** 1% PTCGA200 top-1 accuracies (in %) fine-tuned from supervised self-supervised pretraining on **PTCGA200** and general images.

| Model | Test accuracy | Image size | Pretraining dataset |
|---|---|---|---|
| ResNet50 | 75.64 | 384 | imagenet2012 |
| ResNet50-MoCov2 | 76.08 | 384 | imagenet2012 |
| ResNet50-MoCov2 | 77.88 | 384 | PTCGA200 |
| ResNet50-SimCLR | **82.02** | 384 | PTCGA200 |

| ResNet50-BYOL | 78.82 | 384 | PTCGA200 |
| ViT-S/16 | 78.07 | 384 | in21k |
| ViT-S/16-DINO | 77.59 | 384 | PTCGA200 |
| BiT-S-R50x1 | 77.69 | 384 | imagenet2012 |

We observed the accuracies dropped from 100% **PTCGA200** counterparts as expected. ResNet50-SimCLR fine-tuned on 384px images obtained the best accuracy of 82.02%. Self-supervised ResNet50 with MoCov2, SimCLR and BYOL, and ViT-S/16 with DINO exceeded imagenet2012-pretrained ResNet50 accuracy of 75.64%.

### 6.7 Tiny-PTCGA200 for small experiments.

The original PTCGA200 is imbalanced and too large for experiments requiring many repetitions. We compiled a subset of PTCGA200 to make a small class-balanced dataset, named **tiny-PTCGA200**. **Tiny-PTCGA200** contains randomly resampled 20 patches per slide. 500 slides from each of 6 organs (brain, breast, uterine corpus, kidney, lung, and thyroid) were enrolled. As a use case, we conducted FixMatch [36] and a supervised learning baseline. FixMatch is a semi-supervised learning method matching predictions of strongly-augmented images to weekly-augmented images, working when a small proportion of the data is labeled. We fine-tuned imagenet2012-pretrained EfficientNet-b3 with Adam optimizer with beta1=0.9 and beta2=0.999 without weight decay for 100 epochs. The labeled and unlabeled batch sizes were 16 and 48, respectively. The initial learning rate was 0.001 and decayed with the cosine annealing schedule. We increased the labeled training data (from 1% to 10%) and observed FixMatch surpassed the supervised baseline when the labeled data is below 10% (Fig. 7). This result is in line with the original result on general images [36], underpinning benchmarking utility of **tiny-PTCGA200**.

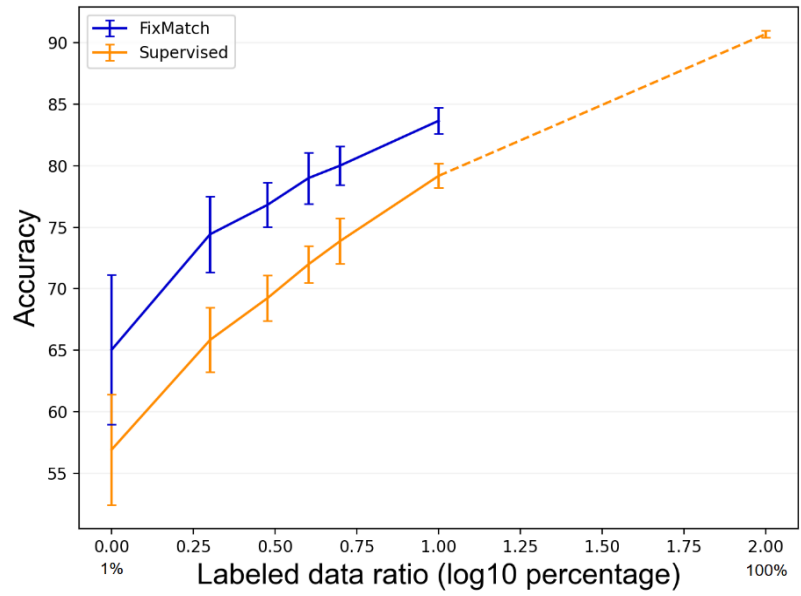

Fig. **6.** Accuracies in % of the FixMatch and EfficientNet-b3 supervised baseline fine-tuned on **tiny-PTCGA200**. 5-fold cross-validation was repeated 3 times for the supervised baseline using the 100% labeled dataset and 8 times for others. The means of the repetition are reported with the ranges shown as error bars.